\documentclass[letter, 10 pt, conference]{ieeeconf}  

\IEEEoverridecommandlockouts                         

\usepackage{hyperref}
\hypersetup{hidelinks,
	colorlinks=true,
	allcolors=red,
	pdfstartview=Fit,
	breaklinks=true}
\usepackage{graphics} 
\usepackage{graphicx}
\usepackage{epsfig} 
\usepackage[linesnumbered,ruled]{algorithm2e}
\usepackage{subcaption}
\usepackage{float}
\usepackage{graphicx}
\usepackage{bm}
\usepackage{times}
\usepackage{balance}
\usepackage{amsmath}
\usepackage{breqn}
\usepackage{cite}
\usepackage{threeparttable}
\usepackage{amssymb}
\usepackage{multirow}
\usepackage{booktabs}
\usepackage{bigstrut}
\usepackage{amsmath}
\usepackage{makecell}
\usepackage{color,soul}

\title{\LARGE \bf
    PlantStereo: A Stereo Matching Benchmark for Plant Surface Dense Reconstruction
}

\author{Qingyu Wang$^{1}$ Baojian Ma$^{1}$ Wei Liu$^{1}$ Mingzhao Lou$^{1}$ Mingchuan Zhou$^{1*}$ Huanyu Jiang$^{1}$ and Yibin Ying$^{1}$ 
	\thanks{${^{1}}$Authors are with College of Biosystems Engineering and Food Science, Zhejiang University, Hangzhou, China.
		{\tt\small \{12013027, mczhou, hyjiang, ybying\}@zju.edu.cn}}%
}%

\begin{document}

	\maketitle
	\thispagestyle{empty}
	\pagestyle{empty}

	\begin{abstract}
	    Stereo matching is an important task in computer vision which has drawn tremendous research attention for decades.
	    While in terms of disparity accuracy, density and data size, public stereo datasets are difficult to meet the requirements of models.
		In this paper, we aim to address the issue between datasets and models and propose a large scale stereo dataset with high accuracy disparity ground truth named PlantStereo.
		We used a semi-automatic way to construct the dataset: after camera calibration and image registration, high accuracy disparity images can be obtained from the depth images.
		In total, PlantStereo contains 812 image pairs covering a diverse set of plants: spinach, tomato, pepper and pumpkin.
		We firstly evaluated our PlantStereo dataset on four different stereo matching methods.
		Extensive experiments on different models and plants show that compared with ground truth in integer accuracy, high accuracy disparity images provided by PlantStereo can remarkably improve the training effect of deep learning models.
		This paper provided a feasible and reliable method to realize plant surface dense reconstruction.
		The PlantStereo dataset and relative code are available at: 
	\end{abstract}
	\href{https://www.github.com/wangqingyu985/}{https://www.github.com/wangqingyu985/PlantStereo}
	
	\section{INTRODUCTION}
	
	Depth estimation is one of the critical topics in 3D reconstruction, phenotyping and breeding automation for crops and plants, which is important to increase food production and solve global famine problem.
	Stereo matching is one of the most fundamental tasks in computer vision.
	Compared with other depth perception systems, the binocular camera system can provide the affordable dense depth estimation, and has been widely applied in autonomous driving\cite{geiger2012we, menze2015object, yang2019drivingstereo, huang2019apolloscape, chang2019argoverse, cordts2016cityscapes} and indoor/outdoor environment depth estimation\cite{scharstein2002taxonomy, scharstein2003high, scharstein2007learning, hirschmuller2007evaluation, scharstein2014high, schops2017multi, bao2020instereo2k, wang2019irs, treible2017cats}.
	However, due to the lack of specialized stereo dataset, there has been little research in plant surface reconstruction using binocular camera system.
	
	In recent years, deep learning methods\cite{kendall2017end, chang2018pyramid, guo2019group, li2021revisiting} greatly improved the performance of stereo matching, bringing in more accurate, faster and more dense depth estimation.
	While deep learning models are data-thirsty, which requires massive labeled disparity images to reach good performance.
	Different from other tasks in computer vision such as image classification, object detection and semantic/instance segmentation, the labeled disparity images in stereo matching task are difficult to obtain in real scene, due to the amount of human labor involved in constructing the scenes and annotating ground truth information\cite{scharstein2002taxonomy}.
	Moreover, the model trained on large-scale synthetic stereo datasets\cite{mayer2016large, butler2012naturalistic} is difficult to generalize in real scene.
	On the other hand, the public stereo datasets nowadays are difficult to meet the requirements of deep learning models in terms of disparity density\cite{yang2019drivingstereo, chang2019argoverse, treible2017cats, geiger2012we, menze2015object} and data size\cite{schops2017multi, ladicky2012joint, scharstein2002taxonomy, scharstein2003high, scharstein2007learning, hirschmuller2007evaluation, scharstein2014high}.
	
	\begin{figure}
		\vspace{0.13cm}
		\centering
		\addtolength{\tabcolsep}{-5pt}
		\renewcommand{\arraystretch}{0.1}
		\begin{tabular}{cc}
            \includegraphics[width=1.0\linewidth]{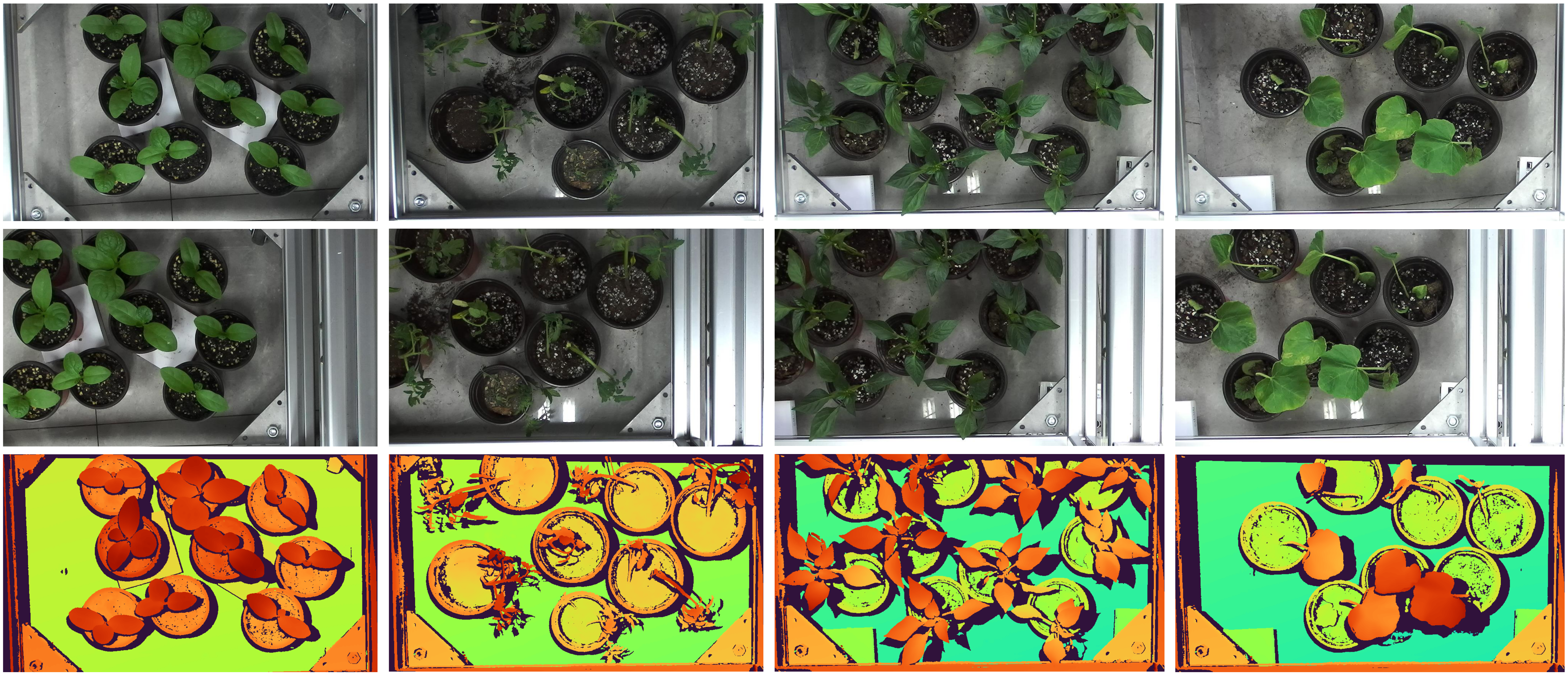}
	        \vspace{0.15cm}\\
		\end{tabular}
		\caption{Some example in PlantStereo dataset, including left image(first row), right image(center row) and disparity image(bottom row), note that the disparity images have been normalized and visualized for demonstration. Best viewed in color.}
		\label{fig:dataset_example}
	\end{figure}
	
	We also noticed that disparity accuracy and data quality of ground truth is another important factor to influence the training effect of deep learning model.
	Before the appearance of deep learning methods, traditional algorithms\cite{hirschmuller2007stereo, kolmogorov2001computing, klaus2006segment} serve stereo matching task as a classification problem, and can only achieve pixel level matching accuracy.
	The emergency of deep learning has brought a revolutionary change to stereo matching task, which define a loss function and convert the original classification problem to regression problem\cite{kendall2017end}.
	Nowadays, the End-Point-Error(EPE) of deep learning models\cite{chang2018pyramid, guo2019group, li2021revisiting} has been less than one on the most popular benchmarks, such as Middlebury\cite{scharstein2002taxonomy, scharstein2003high, scharstein2007learning, hirschmuller2007evaluation, scharstein2014high} and KITTI\cite{geiger2012we, menze2015object}.
	While the most popular datasets\cite{geiger2012we, menze2015object, yang2019drivingstereo, huang2019apolloscape, chang2019argoverse, butler2012naturalistic} are still remaining pixel-level disparity accuracy, which seriously influence the development of deep learning models.
	
	In this article, we aim to address the issue of stereo dataset mentioned above.
	Overall, the main contributions of this paper are listed as follows:
	
	\begin{itemize}
		\item We constructed a data collecting platform equipped with a depth camera and a binocular camera. Based on camera calibration and image registration, we proposed a semi-automatic method which can be used to construct stereo matching dataset.
		\item We developed a stereo matching dataset named \textbf{PlantStereo}, with sub-pixel accuracy disparity map as ground truth, which can be applied to plant surface dense reconstruction.
		\item We demonstrated that our sub-pixel level accuracy disparity ground truth can remarkably improve the training effect of learning-based models, compared with pixel level accuracy ground truth. The proposed method in this paper provides a low-cost solution for plant surface dense  reconstruction.
	\end{itemize}
	
	The rest of this paper is organized as follows:
	in section \ref{sec:2}, we briefly review the existing large scale datasets published for stereo matching task and popular stereo matching algorithms.
	Section \ref{sec:3} introduces the method we proposed to obtain the high accuracy disparity image as ground truth.
	Further more, in section \ref{sec:4}, we will analysis our PlantStereo dataset in detail.
	Section \ref{sec:5} reports our experiments on PlantStereo with different accuracy of ground truth.
	Section \ref{sec:6} concludes the paper and points out the future work.

	\section{RELATED WORK}
	\label{sec:2}
	
	We review the previous works from three primary related streams of the research area: open access stereo matching datasets, traditional stereo matching algorithms and recently developed end-to-end learning based stereo matching algorithms.
	
	\subsection{Stereo Matching Dataset}
	
	In recent years, large-scale datasets applied to stereo matching have been constructed.
	To the best of our knowledge, stereo matching datasets could be divided into six classes according to the tools and methods used to obtain the ground truth disparity image, which are: simulation software, structured light, LIDAR, stereo matching algorithms, annotation and some unsupervised dataset.
	
	\textbf{Synthetic datasets.}
	Simulation software such as UE and Blender are the most widely adopted tools, which allow researchers to build their own scenes, with changeable textures, lighting and weather conditions\cite{he2021semi}.
    Representative datasets constructed using this method include Sintel\cite{butler2012naturalistic}, Scene Flow\cite{mayer2016large}, HR-VS\cite{yang2019hierarchical}, IRS\cite{wang2019irs} and New Tsukuba CG\cite{peris2012towards}.
    In general, disparity images generated by simulation software have the advantages of high accuracy, high density (usually no invalid pixels), and convenient to build large scale datasets.
    However, unfortunately, the generalization performance of the model trained in virtual scenes is unstable when it comes to real scenes, and the real scene datasets are commonly required to fine-tune the model.

	\textbf{Structured light to generate ground truth.}
	Datasets such as Middlebury\cite{scharstein2003high, scharstein2007learning, hirschmuller2007evaluation, scharstein2014high} and InStereo2K\cite{bao2020instereo2k} that use structured light encoding and decoding to get the depth value of the corresponding pixel and then calculate the disparity value.
	The disparity accuracy and density of the dataset constructed by this method can usually meet the requirements of deep learning algorithms, while the encoding and decoding process is complicated and time-consuming, thus it is difficult to construct large scale datasets.

	\textbf{LIDAR to generate ground truth.}
	Representative datasets that use LIDAR or 3D scanner to obtain disparity image include KITTI\cite{menze2015object, geiger2012we}, ETH3D\cite{schops2017multi}, DrivingStereo\cite{yang2019drivingstereo}, ApolloScape\cite{huang2019apolloscape} and Argoverse\cite{chang2019argoverse}.
	Due to the working principle of LIDAR, the ground truth obtained in this way usually have the disadvantage of low disparity density.
	In general, this method requires joint calibration of camera and LIDAR, which is generally applicable to scenes where the object is far from the camera, such as outdoor or autonomous driving.

	\textbf{Stereo matching algorithms to generate ground truth.}
	In addition to the three commonly used methods mentioned above, some researchers also use existing stereo matching algorithms such as SGM\cite{hirschmuller2007stereo} to generate disparity maps.
	Representative datasets include Cityscapes\cite{cordts2016cityscapes}.
	Obviously, the ground truth generated by this method has the shortcomings of low accuracy and low density, thus it has not been widely adopted.
	
	\textbf{Annotation to generate ground truth.}
	In the early days, Middlebury2001\cite{scharstein2002taxonomy} and Ladicky\cite{ladicky2012joint} were constructed by annotating with human labor, with only 8 and 70 pairs of images were provided, respectively.
	The data size is difficult to meet the needs of training a learning based algorithm, thus it is no longer the mainstream method.
	
	\textbf{Unsupervised datasets.}
	In addition, there are some datasets for unsupervised stereo matching, which only provide left and right view images and no ground truth, representative datasets include Flickr1024\cite{wang2019flickr1024}, Holopix50k\cite{hua2020holopix50k} and WSVD\cite{borstler2016role}.
	
\begin{figure*}
		\vspace{0.13cm}
		\centering
		\addtolength{\tabcolsep}{-5pt}
		\renewcommand{\arraystretch}{0.1}
		\begin{tabular}{cc}
            \includegraphics[width=1.0\linewidth]{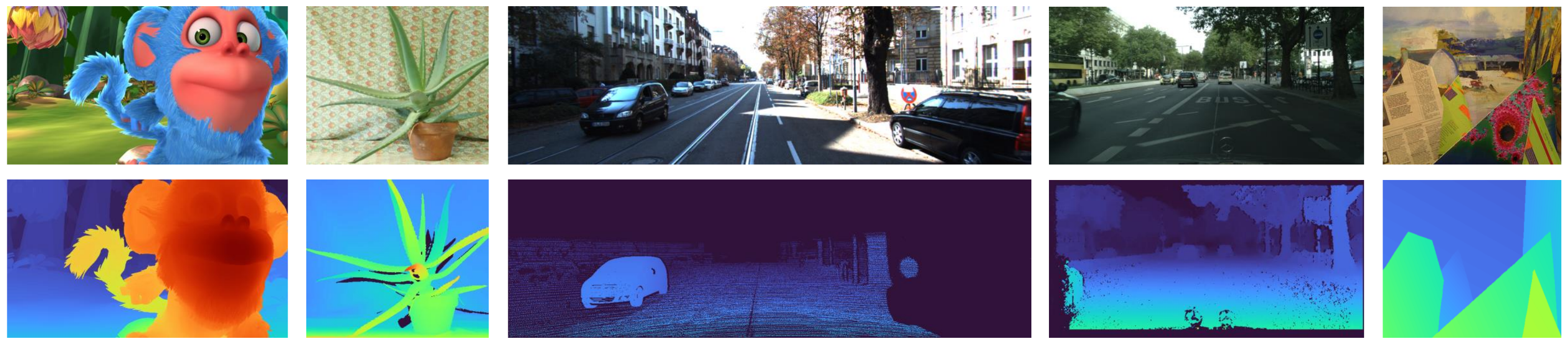}
	        \vspace{0.15cm}\\
		\end{tabular}
		\caption{Representative stereo matching datasets constructed by five methods mentioned above: synthetic dataset(Scene Flow), structured light to generate ground truth(Middlebury2006), LIDAR to generate ground truth(KITTI2015), stereo matching algorithms to generate ground truth(Cityscapes), annotation to generate ground truth(Middlebury2001). The first row represent the left image of the corresponding dataset, the second row represent the disparity images, which have been normalized and visualized for demonstration. Best viewed in color.}
		\label{fig:public_dataset}
	\end{figure*}

	In figure ~\ref{fig:public_dataset}, we give an example of the left image and the associated disparity image for each method to obtain the  ground truth.
	
	\subsection{Traditional Stereo Matching Algorithm}
	
	Traditional stereo matching algorithm calculates the similarity of corresponding image blocks in the left and right images to match.
	In 2002, Scharstein et al. summarized the overall framework of traditional matching algorithm as the following four steps: matching cost computation, cost aggregation, disparity computation and disparity refinement\cite{scharstein2002taxonomy}.
	The traditional stereo matching algorithm can be divided into local matching algorithm, semi-global matching algorithm and global matching algorithm according to the scope of constraints.
	Local matching algorithm uses different cost functions for block matching.
	Semi-global matching algorithm introduces a cost aggregation step to minimize the energy function\cite{hirschmuller2007stereo}.
	Global matching algorithm uses global optimization theory to estimate disparity map, and the representative algorithms include graph cut\cite{kolmogorov2001computing} and belief propagation\cite{klaus2006segment}.
	
	\subsection{Learning Based Stereo Matching Algorithm}
	
	The past 5 years have witnessed the rapid development of deep learning technology applicable to stereo matching task.
	Before the emergence of end-to-end pipeline, researchers have tried to replace local modules with learnable methods\cite{poggi2021synergies}, such as learning the matching cost of image blocks\cite{zbontar2015computing}, disparity optimization\cite{seki2017sgm} and disparity refinement\cite{gidaris2017detect}.
	These local substitutions have greatly improved the accuracy compared with traditional algorithms.
    In recent years, end-to-end pipeline has become the mainstream of research.
    GC-Net\cite{kendall2017end} transforms disparity calculation from a classification problem to a regression problem for the first time, and laying the basic architecture of end-to-end pipeline. PSMNet\cite{chang2018pyramid} uses SPP module and stacked hourglass 3D convolution to carry out image feature extraction and cost aggregation steps, respectively.
    GwcNet\cite{guo2019group} uses grouping method to construct cost volume, which also improved accuracy.
    We also noticed that transformer encoder structure was recently applied to stereo matching task or depth prediction task\cite{ranftl2021vision} and STTR\cite{li2021revisiting} was constructed.
    In addition, there are also researchers adopted multi-task learning: semantic information\cite{wu2019semantic}, normal vector extraction\cite{kusupati2020normal}, edge extraction\cite{song2018edgestereo} and others on stereo matching task.

	\section{METHOD}
	\label{sec:3}
	In this section, we will briefly introduce the camera calibration and image registration method we proposed to obtain ground truth disparity image.

	\subsection{Sensor Platform}
	
	In this research, we adopted two cameras to construct PlantStereo dataset: mech to obtain depth information and serve as ground truth disparity, and ZED to capture left and right image pairs.
	The camera parameters such as Field of View(FoV), resolution, working range and depth accuracy adopted in this research in detail are listed in table \ref{tab:sensor_platform}.
	
\begin{table*}[htbp]
  \centering
  \caption{Camera Parameter Employed in Our Research}
    \begin{tabular}{cccccc}
    \toprule
    \textbf{Camera} & \textbf{Type} & \textbf{Field of View(°)} & \textbf{Resolution} & \textbf{Range(m)} & \textbf{Accuracy} \\
    \midrule
    Mech-Mind Pro S Enhanced & structured light depth camera & 38.6×24.8 & 1920×1200 & 0.5-1.0 & 0.05mm@0.6m \\
    Stereolabs ZED & stereo camera & 90×60 & 2208×1242 & 0.3-25 & / \\
    \bottomrule
    \end{tabular}%
  \label{tab:sensor_platform}%
\end{table*}%
	
\begin{figure}
		\vspace{0.13cm}
		\centering
		\addtolength{\tabcolsep}{-5pt}
		\renewcommand{\arraystretch}{0.1}
		\begin{tabular}{cc}
            \includegraphics[width=1.0\linewidth]{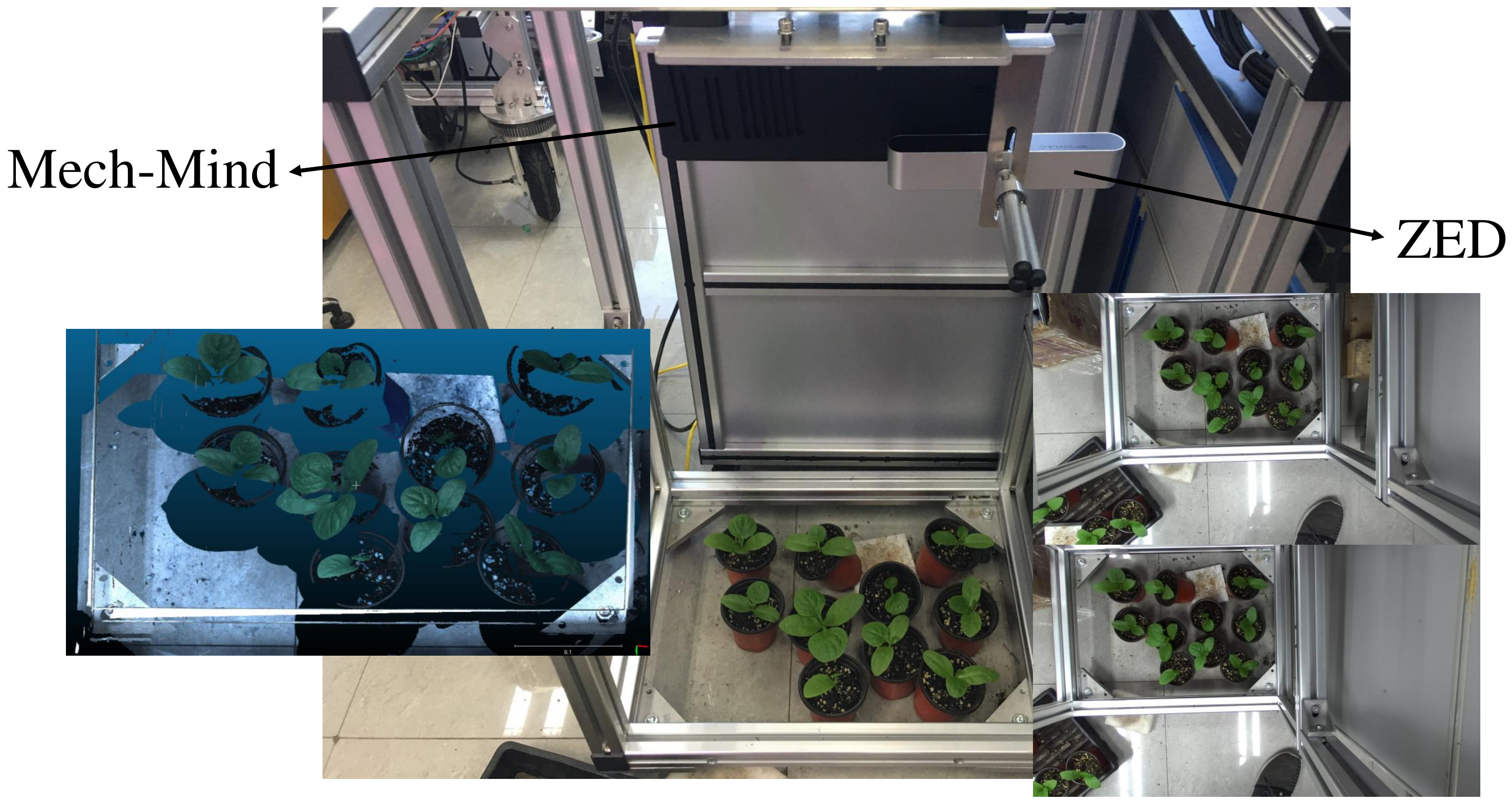}
	        \vspace{0.15cm}\\
		\end{tabular}
		\caption{Data sampling system constructed in this research.}
		\label{fig:sampling_system}
	\end{figure}

	We built an image acquisition system to construct datasets, as shown in Figure~\ref{fig:sampling_system}.
    During the experiment to construct the dataset, the relative position of the two cameras is fixed through a specially designed fastenings.
	The rotation and translation matrix between two cameras are calculated through camera calibration step.
	
	\subsection{Camera Calibration}
	
\begin{figure}
		\vspace{0.13cm}
		\centering
		\addtolength{\tabcolsep}{-5pt}
		\renewcommand{\arraystretch}{0.1}
		\begin{tabular}{cc}
            \includegraphics[width=0.7\linewidth]{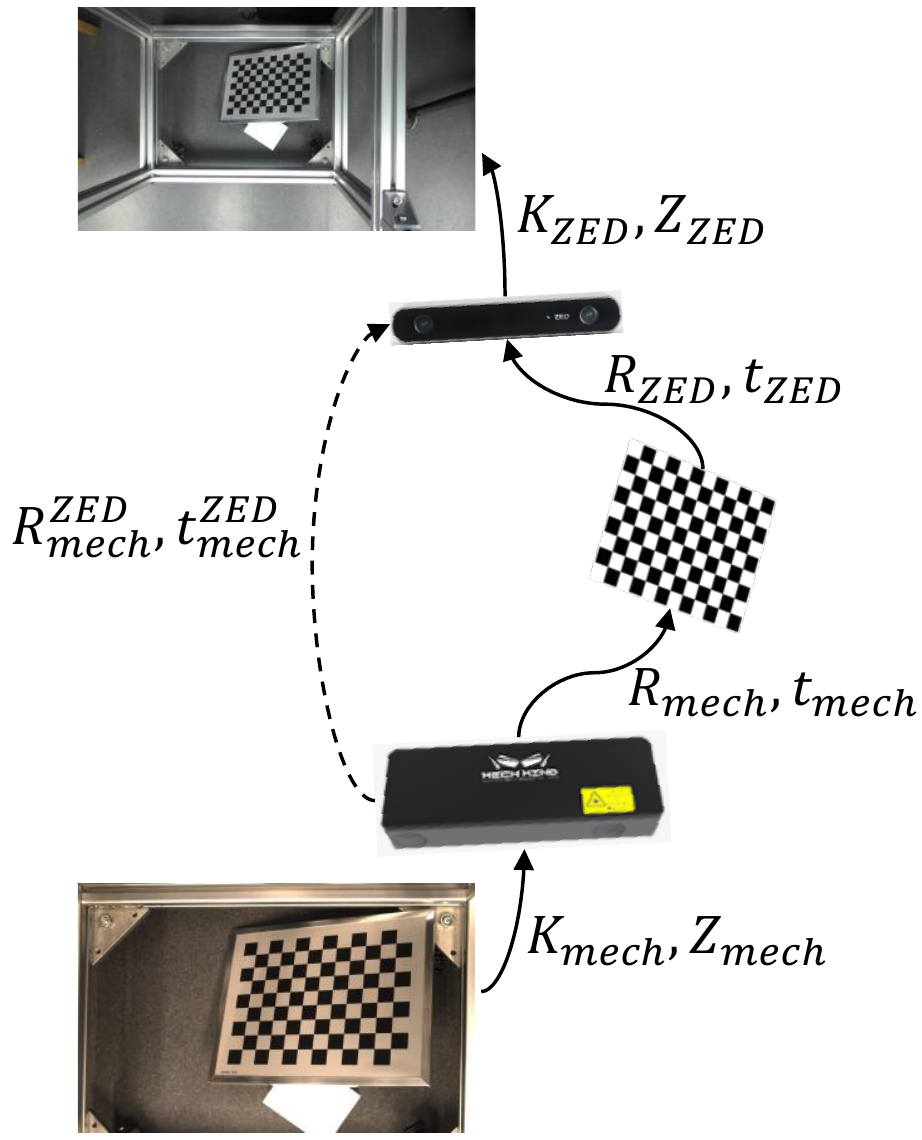}
	        \vspace{0.15cm}\\
		\end{tabular}
		\caption{The schematic diagram representing the method for coordinate transformation from mech camera to ZED camera.}
		\label{fig:trmatrix}
	\end{figure}

	Zhang's monocular camera calibration method\cite{zhang2000flexible} can only obtain the transformation matrix between the camera and the world coordinate system.
	By considering the world coordinate system as the interchange coordinate, we can calculate the rotation and translation matrix between the mech camera and the ZED camera through equation \ref{eq:rtmatrix}, as illustrated in Figure~\ref{fig:trmatrix}.
	
	\begin{equation}\label{eq:rtmatrix}
        \begin{cases}
            R_{mech}^{ZED}=R_{ZED}R_{mech}^{-1}  
            \\
            t_{mech}^{ZED}=t_{ZED}-R_{ZED}R_{mech}^{-1}t_{mech}
        \end{cases}
	\end{equation}
	where $R_{mech}$ and $R_{ZED}$ denote the rotation matrix of the mech camera and the ZED camera, respectively.
	Accordingly, $t_{mech}$ and $t_{ZED}$ denotes the translation matrix of the mech camera and the ZED camera, respectively.
	The above extrinsic matrix could be obtained through Zhang's camera calibration method\cite{zhang2000flexible}.
	In order to make the calibration result more accurate, we calibrated the two cameras for several times and averaged the results.
	
	\subsection{Image Registration}
	
	We generate the ground truth disparity image by registering the depth image captured by mech camera on the left image captured by ZED camera.
	
	First, the pixel coordinate on mech image $I_{i,mech}$ was transformed to world coordinate from the view of mech camera $P_{i,mech}$ throuth equation \ref{eq:mechpixel2mechworld}.
	
	\begin{equation}\label{eq:mechpixel2mechworld}
        P_{i,mech}=\left ( K_{mech}  \right ) ^{-1} \times I_{i,mech} \times Z_{mech}
	\end{equation}
	
	Where $K_{mech}$ and $Z_{mech}$ denote the intrinsic matrix of the mech camera and the depth value of the exact pixel, respectively.
	Then, the world coordinate was transformed from the view of the mech camera to the view of the ZED camera by means of the mutual rotation and translation matrix($R_{mech}^{ZED}$ and $t_{mech}^{ZED}$) obtained from equation \ref{eq:rtmatrix}.
	
	\begin{equation}\label{eq:mechworld2zedworld}
        P_{i,ZED}=R_{mech}^{ZED}\times P_{i,mech} + t_{mech}^{ZED}  
	\end{equation}
	
	Finally, the point was transformed back to pixel coordinate of the ZED camera $I_{i,ZED}$.
	
	\begin{equation}\label{eq:zedworld2zedpixel}
        I_{i,ZED}=\frac{K_{ZED}\times P_{i,ZED}  }{Z_{ZED} }
	\end{equation}
	
	where $K_{ZED}$ is the intrinsic matrix of the ZED camera, while depth was indicated as $Z_{ZED}$, which equals to the value of the third element of the $P_{i,ZED}$ calculated from equation \ref{eq:rtmatrix}.
	
	Thus, each pixel in the image captured by mech camera could be aligned to ZED image through equation \ref{eq:mechpixel2mechworld}, \ref{eq:mechworld2zedworld} and \ref{eq:zedworld2zedpixel}.
	After transform the depth to disparity on each pixel through equation \ref{eq:depth2disparity}, a disparity image with high accuracy and density could be generated.
	
    \begin{equation}\label{eq:depth2disparity}
        d_{i}=\frac{b_{ZED}\times f_{ZED}}{Z_{ZED}} 
	\end{equation}
	
	where $b_{ZED}$ is the baseline of the ZED camera and $f_{ZED}$ is the focal length of the ZED camera.
	Both intrinsic parameters could be obtained through the camera calibration step\cite{zhang2000flexible}.
	
    \section{ANALYSIS}
	\label{sec:4}
	
	\subsection{Dataset Overview}
    \label{sec:4.1}
	
	A total of 812 pairs of left and right images with high accuracy disparity images were collected to construct the PlantStereo dataset.
	Among them, 470 pairs of images were used for training, 110 pairs were used to validate the training effect, and another 232 pairs were used to test the final training results.
	Besides, 300 pairs of images were captured for spinach reconstruction, 150 for tomato, 212 for pepper and another 150 for pumpkin, more details about PlantStereo are illustrated in Table\ref{tab:dataset_information}.
	
\begin{table}[htbp]
  \centering
  \caption{Basic Information of the PlantStereo Dataset}
    \begin{tabular}{cccccc}
    \toprule
    \textbf{Subset} & \textbf{Train} & \textbf{Validation} & \textbf{Test} & \textbf{All} & \textbf{Resolution} \\
    \midrule
    \textbf{Spinach} & 160   & 40    & 100   & 300   & 1046×606 \\
    \textbf{Tomato} & 80    & 20    & 50    & 150   & 1040×603 \\
    \textbf{Pepper} & 150   & 30    & 32    & 212   & 1024×571 \\
    \textbf{Pumpkin} & 80    & 20    & 50    & 150   & 1024×571 \\
    \midrule
    \textbf{All} & 470   & 110    & 232   & 812   &  \\
    \bottomrule
    \end{tabular}%
  \label{tab:dataset_information}%
\end{table}%

    For the ground truth of disparity images at pixel level accuracy, we save them as normal 8 bit \textbf{.png} format.
    For disparity images at sub-pixel accuracy, we save the 32 bit floating data as \textbf{.tiff} format. 
    
    \subsection{Dataset Comparison}
    \label{sec:4.2}
    
    By comparing with the existing public datasets, we found that our PlantStereo dataset showed superiority in many aspects.
	
    To the author's knowledge, PlantStereo is the first stereo matching dataset applied in agriculture, and constructed in real scenario, which can promote the generalization performance of the deep learning model in practical application.
    
    In terms of data size, PlantStereo exceeds the earlier datasets and is sufficient for training the deep learning model.
    
    In terms of the accuracy of the disparity map, only three datasets: Middlebury2014\cite{scharstein2002taxonomy}, HR-VS\cite{yang2019hierarchical} and IRS\cite{wang2019irs} achieved sub-pixel accuracy, and the latter two\cite{yang2019hierarchical, wang2019irs} were synthetic datasets generated by software.
    Obviously, as the average matching error of the SOTA deep learning stereo matching model has been less than one pixel, most datasets that only provide pixel-level accuracy disparity maps can no longer meet the requirements of the deep learning models.
    
    In terms of disparity density, PlantStereo is close to 90\%, which is in the forefront of the existing datasets.
    It is obviously better than the datasets acquired ground truth by LiDAR, and only lower than the datasets generated by simulation software.
    
    We summarize the public stereo matching datasets in table \ref{tab:datasets_comparison} for comparison.
	
\begin{table*}[htbp]
  \centering
  \caption{Comparison between PlantStereo Dataset and the Other Polular Stereo Matching Datasets Published}
    \begin{tabular}{ccccccc}
    \toprule
    \textbf{Dataset} & \textbf{Tools} & \textbf{Scene Type} & \textbf{Data Size} & \textbf{Accuracy} & \textbf{Real} & \textbf{Density} \\
    \midrule
    Middlebury \cite{scharstein2002taxonomy, scharstein2003high, scharstein2007learning, hirschmuller2007evaluation, scharstein2014high} & structured light & indoor & 8+2+9+21+30 & sub-pixel only in \cite{scharstein2014high} & \checkmark     & $\approx$ 96 \% \\
    KITTI \cite{geiger2012we, menze2015object} & LiDAR & autonomous driving & 389+400 & pixel & \checkmark     & $\approx$ 19 \% \\
    Scene Flow \cite{mayer2016large} & software & animation & 39049 & pixel & ×     & 100 \% \\
    HR-VS \cite{yang2019hierarchical}    & software & autonomous driving & 780   & sub-pixel & ×     & 100 \% \\
    New Tsukuba  \cite{peris2012towards}  & software & indoor & 1800  & pixel & ×     & 100 \% \\
    ETH3D \cite{schops2017multi} & scanner & indoor/outdoor & 27(train)+20(test) & pixel & \checkmark     & $\approx$ 69 \% \\
    DrivingStereo \cite{yang2019drivingstereo} & LiDAR & autonomous driving & 174437(train)+7751(test) & pixel & \checkmark     &    $\approx$ 4 \% \\
    InStereo2K  \cite{bao2020instereo2k} & structured light & indoor & 2010(train)+50(test) & pixel & \checkmark     & $\approx$ 87 \% \\
    ApolloScape  \cite{huang2019apolloscape} & LiDAR & autonomous driving & 4156(train)+1009(test) & pixel & \checkmark     &       $\approx$ 90 \% \\
    Argoverse  \cite{chang2019argoverse} & LiDAR & autonomous driving & 4008(train)+1522(validate)+1094(test) & pixel & \checkmark     &  $\approx$ 0.86 \% \\
    Sintel  \cite{butler2012naturalistic} & software & animation & 1064  & pixel & ×     & 100\% \\
    IRS \cite{wang2019irs}   & software & indoor & 84946(train)+15079(test) & sub-pixel & ×     & 100\% \\
    CATS  \cite{treible2017cats}  & LiDAR & indoor/outdoor & 1372  & pixel & \checkmark     &  $\approx$ 8\% \\
    Ladicky  \cite{ladicky2012joint} & annotation & autonomous driving & 50(train)+20(test) & pixel & \checkmark     &  \\
    Cityscapes  \cite{cordts2016cityscapes} & SGM algorithm & autonomous driving &       & pixel & \checkmark     & $\approx$ 38\% \\
    \midrule
    \textbf{PlantStereo} & \textbf{depth camera} & \textbf{agriculture} &   \textbf{470(train)+110(validate)+232(test)}    & \textbf{sub-pixel} & \textbf{\checkmark} & \textbf{$\approx$ 88\%}  \\
    \bottomrule
    \end{tabular}%
  \label{tab:datasets_comparison}%
\end{table*}%
    
\begin{figure}
		\vspace{0.13cm}
		\centering
		\addtolength{\tabcolsep}{-5pt}
		\renewcommand{\arraystretch}{0.1}
		\begin{tabular}{cc}
            \includegraphics[width=0.9\linewidth]{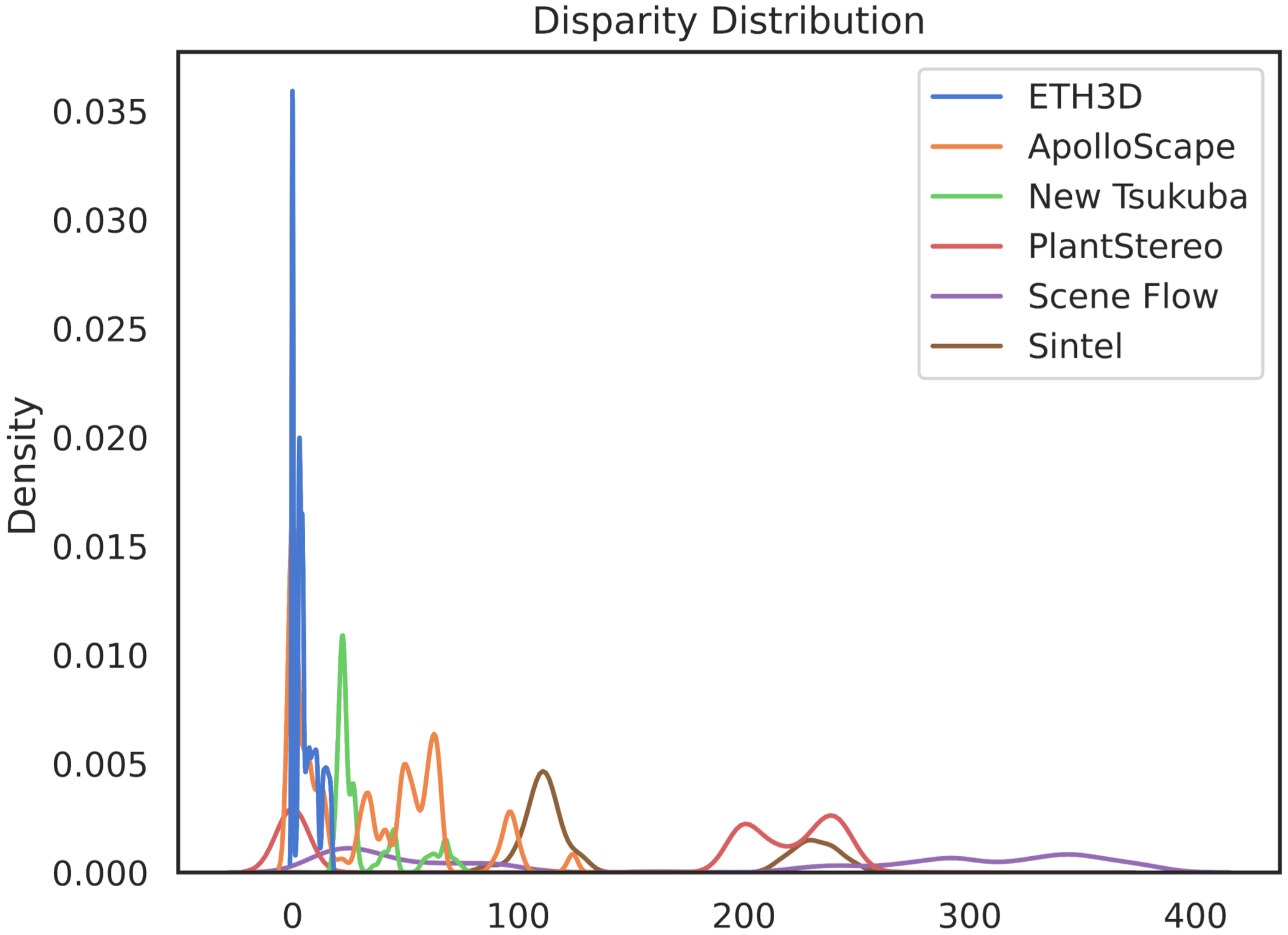}
	        \vspace{0.15cm}\\
		\end{tabular}
		\caption{Disparity distribution of different datasets: Scene Flow\cite{mayer2016large}, ApolloScape\cite{huang2019apolloscape}, Sintel\cite{butler2012naturalistic}, New Tsukuba\cite{peris2012towards}, ETH3D\cite{schops2017multi} and PlantStereo (our dataset).}
		\label{fig:disparity_distribution}
	\end{figure}

    Besides, we evaluated the disparity distribution of PlantStereo and other datasets in Figure~\ref{fig:disparity_distribution}.
    Obviously, the disparity distribution histogram of PlantStereo is bimodal, because the ground and leaf surface occupy most of the pixels.
    In addition, different from other datasets with disparity distribution in $\left [ 0,D_{max}  \right ]$ , PlantStereo's disparity is ranging from 200 to 260, and the minimum disparity $D_{min}$ is not 0.
    This is because the furthest distance in the image is ground, rather than the infinite distance in outdoor scenes such as autonomous driving. 
    Compared with other datasets, the larger maximum disparity $D_{max}$ also increases the difficulty, and can more truly reflect the matching effect of the model in difficult scenes such as large disparity and close distance.
    
    \subsection{Quantitative Analysis}
    \label{sec:4.3}
    
    In order to evaluate the image registration accuracy between disparity image and left image in PlantStereo, quantitative analysis was performed.
    
\begin{figure}
		\vspace{0.13cm}
		\centering
		\addtolength{\tabcolsep}{-5pt}
		\renewcommand{\arraystretch}{0.1}
		\begin{tabular}{cc}
            \includegraphics[width=1.0\linewidth]{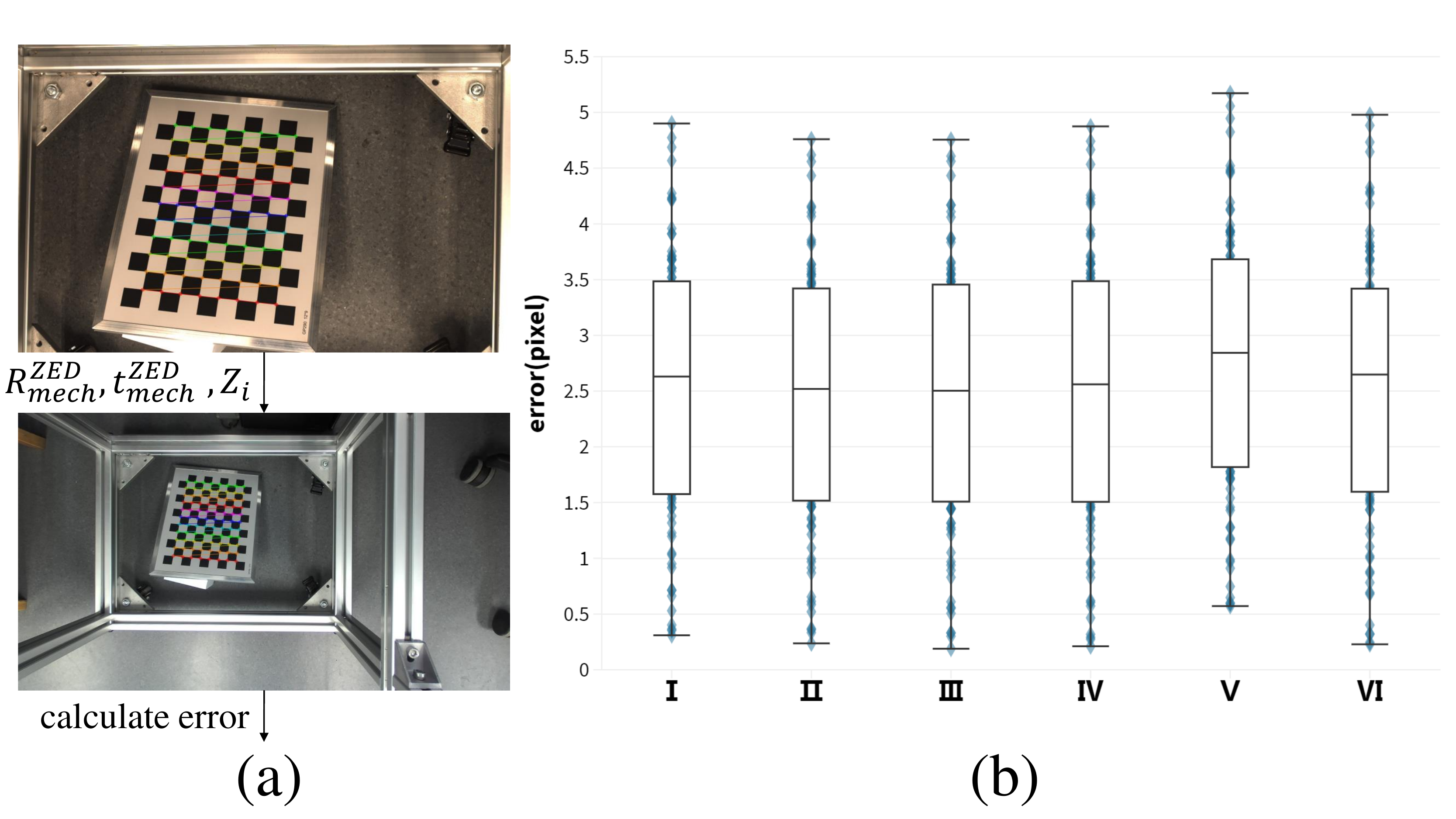}
	        \vspace{0.15cm}\\
		\end{tabular}
		\caption{Quantitative analysis of the image registration error. (a) The schematic of the calculation method. (b) Testing results for six times on 8×11 chessboard inner corners.}
		\label{fig:image_registration_error}
	\end{figure}

    In this step, 8×11 inner corners of the chessboard are used as the testing objects.
    We first detect the corners in the image coordinates of depth camera.
    Then, through $R_{mech}^{ZED}$ and $t_{mech}^{ZED}$ matrix calculated from the camera calibration steps and the depth of inner corner $Z_{i}$ captured by depth camera, we can calculate the image coordinates of the corresponding corner in left image.
    With the image coordinate of the corresponding corner detected by OpenCV in the left image, we can calculate the Euclidean distance between two coordinates (in pixels), as the registration error. The schematic of the image registration error calculation method was shown in Figure~\ref{fig:image_registration_error} (a).
    We implemented six times of testing with an average registration error of 2.5 pixels, as shown in Figure~\ref{fig:image_registration_error} (b).
    
	\section{EXPERIMENTS AND RESULTS}
	\label{sec:5}
	
	In this section, we used four methods: BM, SGM\cite{hirschmuller2007stereo}, PSMNet\cite{chang2018pyramid} and GwcNet\cite{guo2019group} to evaluate on PlantStereo.
	We also performed ablation study to evaluate the influence made by disparity accuracy of ground truth.
	The experimental implementation details are presented in section \ref{sec:5.1}, followed by the evaluation metrics, results and discussion.

    \subsection{Implementation Details}
    \label{sec:5.1}
    
    The two traditional algorithms: BM and SGM\cite{hirschmuller2007stereo} were implemented using python and OpenCV.
    For BM, the block size was set to 15.
    For SGM, the matching block size was set to 3, the penalty coefficient $P_1$ and $P_2$ were set to 216 and 864, respectively. In the  process of left and right consistency check, we set the maximum difference to 1.
    
    The two learning based model: PSMNet\cite{chang2018pyramid} and GwcNet\cite{guo2019group} were implemented using PyTorch architecture.
    both models were end-to-end trained with Adam($\beta_{1}=0.9$, $\beta_{2}=0.999$).
    We performed color normalization (normalize each channel of the image by subtracting their means and dividing their standard deviations) on the entire PlantStereo dataset for data preprocessing.
    The learning rate of the training process began at 0.001 for the first 200 epochs, and 0.0001 for the remaining 300 epochs.
    The batch size was fixed to 1 for the training process on one Nvidia RTX 3090 GPU.

	\subsection{Evaluation Metrics}
	
	In order to evaluate the performance of the algorithms in a quantitative way, we adopted three general methods called bad-$\delta$ error, End-Point-Error(EPE) and Root Mean Square Error(RMSE) to calculate the matching accuracy.
	The calculation method of the three metrics are shown in equation \ref{eq:bad-delta-error}, \ref{eq:end-point-error} and \ref{eq:root mean square error}.
	
	\begin{equation}
	\label{eq:bad-delta-error}
        bad- \delta =\frac{\sum_{\left ( x,y \right ) }^{} \left [ \left | \hat{d} \left ( x,y \right ) - d^{*} \left ( x,y \right )   \right | > \delta  \right ]  }{N} \times 100 \%
	\end{equation}
	
	where $\hat{d} \left ( x,y \right )$ is the disparity calculated by algorithms, $d^{*} \left ( x,y \right )$ is the ground truth disparity value provided by dataset.
	Operator $\left [ \cdot  \right ]$ indicates the value becomes 1 if the condition establish.
	$N$ is the effective pixel number in one disparity image, and effective pixel must meets the requirement that $0 < d^{*}\left ( x,y \right )  < D_{max} $, where $D_{max}$ is the maximum matching disparity given by us.
	
	\begin{equation}
	\label{eq:end-point-error}
        EPE = \frac{\sum_{\left ( x,y \right ) }^{} \left | \hat{d} \left ( x,y \right ) - d^{*} \left ( x,y \right )  \right |  }{N} 
	\end{equation}
	
	where all the terms have the same meaning as equation \ref{eq:bad-delta-error}.
	
	\begin{equation}
	\label{eq:root mean square error}
        RMSE = \sqrt{\frac{\sum_{\left ( x,y \right ) }^{}  \left ( \hat{d} \left ( x,y \right ) - d^{*} \left ( x,y \right )  \right )^{2}    }{N} }
    \end{equation}
	
	where all the terms have the same meaning as equation \ref{eq:end-point-error}.

    \subsection{Computation Volume and Inference Time}
    
    Considering the tradeoff between accuracy and inference time, the maximum matching disparity $D_{max}$ was set to 256 for four  algorithms.
    During training process, all the input images were randomly cropped to size 256$\times$512.
    Under such a condition, we calculated the model parameters (\#param.) and Floating point operations (GFLOPs) for both learning-based methods, as shown in table \ref{tab:computation_volume_and_inference_time}.
    Obviously, in the case of less computing volume, GwcNet possess more learnable parameters to fit a better model.
    
    As for inference process, all the input left and right images were padded with zeros on the top and right sides to become the resolution of 608$\times$1056.
    Under the condition mentioned above, we tested on PlantStereo and found that BM and GwcNet consume 0.02s in average to inference a single pair of image.
    On the other hand, PSMNet consumes 1.05s in average and can not satisfy the real time requirement.
    
    \begin{figure*}
		\vspace{0.13cm}
		\centering
		\addtolength{\tabcolsep}{-5pt}
		\renewcommand{\arraystretch}{0.1}
		\begin{tabular}{cc}
            \includegraphics[width=0.9\linewidth]{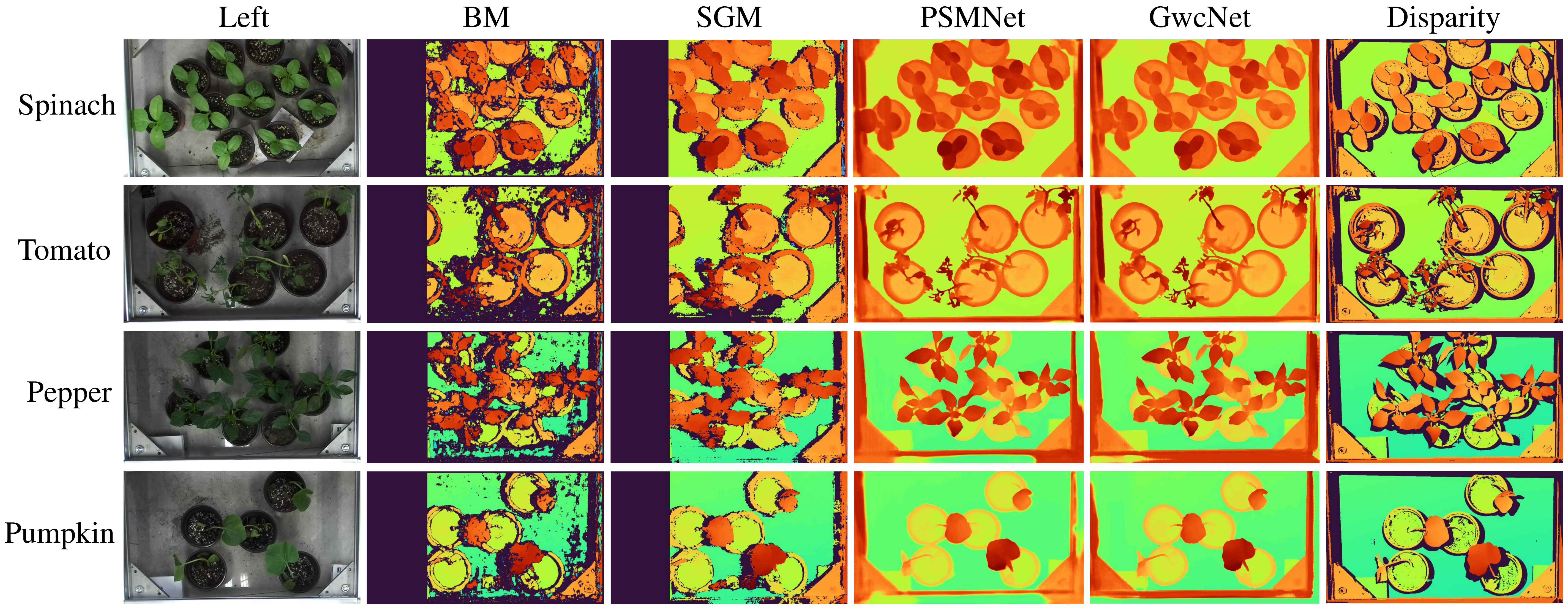}
	        \vspace{0.15cm}\\
		\end{tabular}
		\caption{Results on test set(four methods: BM, SGM\cite{hirschmuller2007stereo}, PSMNet\cite{chang2018pyramid} and GwcNet\cite{guo2019group} on four subset: spinach, tomato, pepper and pumpkin). Best viewed in color.}
		\label{fig:result}
	\end{figure*}
	
\begin{table}[htbp]
  \centering
  \caption{Computation Volume and Inference Time Comparison Between Different Method}
    \begin{tabular}{cccc}
    \toprule
    \textbf{Method} & \textbf{\#param.(M)} & \textbf{GFLOPs} & \textbf{Time(s)} \\
    \midrule
    BM    &   /    &    /   & \textbf{0.02} \\
    SGM   &   /    &    /   & 0.19 \\
    PSMNet & 5.36  & 29.22 & 1.05 \\
    GwcNet & \textbf{6.43} & \textbf{26.13} & \textbf{0.02} \\
    \bottomrule
    \end{tabular}%
  \label{tab:computation_volume_and_inference_time}%
\end{table}%

    \subsection{Matching Accuracy}
    
	In our experiment, the value $\delta $ of bad-$\delta$ error was set to 1, 3 and 5 to evaluate the corresponding proportion of pixels failed to match correctly.
	
	Compared SGM with BM algorithm, bacause of the cost aggregation and the disparity refinement steps, SGM can perform much better than BM on both validation and testing set, especially in textureless regions such as ground and plant surface.
	Due to the perspective transformation, there are some occluded regions which can be seen on left image but can not be seen on right image.
	The traditional algorithms can not give a specific prediction on these misleading regions.
	Thus, the matching accuracy of traditional algorithms is much lower that that of learning based methods, as shown in table \ref{tab:matching_accuracy}.
	
	Compared GwcNet with PSMNet, because of the improvement in cost volume construction and 3D convolution steps, GwcNet can attain better fitting effect.
	And both two end-to-end models can perform much better than traditional algorithms.
    
\begin{table}[htbp]
  \centering
  \caption{Matching Accuracy Comparison Between Different Method on Validation Set}
    \begin{tabular}{cccccc}
    \toprule
    \textbf{Method} & \textbf{bad-1(\%)} & \textbf{bad-3(\%)} & \textbf{bad-5(\%)} & \textbf{EPE} & \textbf{RMSE} \\
    \midrule
    BM    & 85.83 & 50.12 & 49.57 & 102.79 & 147.9 \\
    SGM   & 71.55 & 37.08 & 36.21 & 71.48 & 122.30 \\
    PSMNet & 29.81 & 4.88  & 3.17  & 1.21  & 3.2 \\
    GwcNet & \textbf{18.11} & \textbf{2.9} & \textbf{1.77} & \textbf{0.84} & \textbf{2.56} \\
    \bottomrule
    \end{tabular}%
  \label{tab:matching_accuracy}%
\end{table}%
	
    \subsection{Ablation Study on Disparity Accuracy}
    
    In this section, in order to verify the improvement of training effect brought by the disparity accuracy of ground truth, we performed ablation study on two learning based methods\cite{chang2018pyramid}\cite{guo2019group} and four kinds of plants in PlantStereo.
    The results are shown in table \ref{tab:ablation_disparity_accuracy}, where \textbf{low} represents pixel level accuracy disparity ground truth and \textbf{high} represents sub-pixel level accuracy disparity ground truth.
    
    The results indicate that the performance on both validation set and testing set becomes better with the increment of disparity accuracy from pixel level to sub-pixel level.
    It is worth noting that out high accuracy ground truth improved the performance without increase the parameters or inference time of the model.
    
\begin{table*}[htbp]
  \centering
  \caption{Ablation Study on Disparity Accuracy of Ground Truth.}
    \begin{tabular}{ccc|ccccc|ccccc}
    \toprule
    \multirow{2}[4]{*}{\textbf{Model}} & \multirow{2}[4]{*}{\textbf{Subset}} & \multirow{2}[4]{*}{\textbf{Accuracy}} & \multicolumn{5}{c|}{\textbf{Validation}} & \multicolumn{5}{c}{\textbf{Test}} \\
\cmidrule{4-13}          &       &       & \textbf{bad-1(\%)} & \textbf{bad-3(\%)} & \textbf{bad-5(\%)} & \textbf{EPE} & \textbf{RMSE} & \textbf{bad-1(\%)} & \textbf{bad-3(\%)} & \textbf{bad-5(\%)} & \textbf{EPE} & \textbf{RMSE} \\
    \midrule
    \multirow{8}[8]{*}{PSMNet} & \multirow{2}[2]{*}{spinach} & low   & 29.09 & 3.48  & 2.39  & 1.03  & 2.38  & 46.98 & 8.77  & 5.29  & 1.46  & 2.56 \\
          &       & high  & \textbf{12.09} & \textbf{1.94} & \textbf{1.06} & \textbf{0.63} & \textbf{1.67} & \textbf{33.48} & \textbf{3.4} & \textbf{1.97} & \textbf{1.03} & \textbf{2.13} \\
\cmidrule{2-13}          & \multirow{2}[2]{*}{tomato} & low   & 20.13 & 5.15  & 3.57  & 1.14  & 3.64  & 21.71 & 5.59  & 3.9   & 1.17  & 3.67 \\
          &       & high  & \textbf{17.22} & \textbf{4.23} & \textbf{2.54} & \textbf{0.94} & \textbf{3.22} & \textbf{18.3} & \textbf{4.46} & \textbf{2.71} & \textbf{1.01} & \textbf{3.45} \\
\cmidrule{2-13}          & \multirow{2}[2]{*}{pepper} & low   & 31.97 & 4.01  & 2.39  & 1.2   & 3.46  & 33.33 & 4.66  & 3.08  & 1.38  & 4.11 \\
          &       & high  & \textbf{24.87} & \textbf{3.59} & \textbf{2.33} & \textbf{1.13} & \textbf{3.39} & \textbf{25.7} & \textbf{3.68} & \textbf{2.36} & \textbf{1.15} & \textbf{3.43} \\
\cmidrule{2-13}          & \multirow{2}[2]{*}{pumpkin} & low   & 20.18 & 4.43  & 3.09  & 0.97  & 3.22  & 25.88 & 5.25  & 3.24  & 1.22  & 3.68 \\
          &       & high  & \textbf{15.04} & \textbf{3.21} & \textbf{2.31} & \textbf{0.82} & \textbf{2.9} & \textbf{17.22} & \textbf{4.23} & \textbf{2.54} & \textbf{0.94} & \textbf{2.93} \\
    \midrule
    \multirow{8}[8]{*}{GwcNet} & \multirow{2}[2]{*}{spinach} & low   & 10.61 & 1.71  & 1.3   & 0.54  & 1.86  & 21.5  & 2.04  & 1.22  & 0.83  & 1.87 \\
          &       & high  & \textbf{10.58} & \textbf{1.26} & \textbf{0.73} & \textbf{0.52} & \textbf{1.53} & \textbf{21.42} & \textbf{1.94} & \textbf{1.15} & \textbf{0.83} & \textbf{1.86} \\
\cmidrule{2-13}          & \multirow{2}[2]{*}{tomato} & low   & 15.39 & 3.6   & 2.61  & 0.92  & 3.34  & 18.12 & 4.69  & 3.12  & 1.06  & 3.57 \\
          &       & high  & \textbf{13.94} & \textbf{3.48} & \textbf{2.33} & \textbf{0.87} & \textbf{3.25} & \textbf{14.96} & \textbf{4.03} & \textbf{2.81} & \textbf{0.91} & \textbf{3.29} \\
\cmidrule{2-13}          & \multirow{2}[2]{*}{pepper} & low   & 13.63 & 2.87  & 1.96  & 0.79  & 3.13  & 14.82 & 3.54  & 2.73  & 0.84  & 3.43 \\
          &       & high  & \textbf{12.48} & \textbf{2.57} & \textbf{1.71} & \textbf{0.76} & \textbf{3.09} & \textbf{13.03} & \textbf{2.68} & \textbf{1.78} & \textbf{0.78} & \textbf{3.21} \\
\cmidrule{2-13}          & \multirow{2}[2]{*}{pumpkin} & low   & 15.04 & 3.42  & 2.31  & 0.82  & 2.9   & 15.51 & 3.75  & 2.7   & 0.9   & 3.18 \\
          &       & high  & \textbf{14.55} & \textbf{3.21} & \textbf{2.07} & \textbf{0.72} & \textbf{2.58} & \textbf{14.63} & \textbf{3.71} & \textbf{2.45} & \textbf{0.79} & \textbf{2.89} \\
    \bottomrule
    \end{tabular}%
  \label{tab:ablation_disparity_accuracy}%
\end{table*}%

	\section{CONCLUSION AND FUTURE WORK}
	\label{sec:6}
	
	\subsection{Conclusion}
	
	In this work, we introduced a new stereo matching benchmark with high accuracy disparity image, PlantStereo for plant surface dense reconstruction.
	Left image, right image and high precision disparity image generated from depth image of four kinds of plants collected in real scene are concluded in the dataset.
	We demonstrated that our sub-pixel level accuracy disparity ground truth can remarkably improve the training effect of learning-based models, compared with pixel level accuracy ground truth.

    We have released the dataset, sampling and testing code publicly so that the community can research based on our work.

	\subsection{Future Work}
	
	Due to the characteristics of PlantStereo introduced in section \ref{sec:4}, potential research directions include:
	\begin{itemize}
		\item Continuous breakthrough on the existing sub-pixel level matching accuracy.
		\item Break the limit of maximum disparity $D_{max}$ as a necessary parameter of the model.
		\item Achieve real-time matching under the consumption level computing power.
	\end{itemize}
	
	We expect the newly released PlantStereo benchmark can inspire researchers to develop new and more robust stereo matching algorithms.
	
	\bigskip
	
	\section*{ACKNOWLEDGMENT}
	We gratefully acknowledge the funding.

	\bibliographystyle{IEEEtran}
	\bibliography{my_cite}

\begin{thebibliography}{10}
\providecommand{\url}[1]{#1}
\csname url@rmstyle\endcsname
\providecommand{\newblock}{\relax}
\providecommand{\bibinfo}[2]{#2}
\providecommand\BIBentrySTDinterwordspacing{\spaceskip=0pt\relax}
\providecommand\BIBentryALTinterwordstretchfactor{4}
\providecommand\BIBentryALTinterwordspacing{\spaceskip=\fontdimen2\font plus
\BIBentryALTinterwordstretchfactor\fontdimen3\font minus
  \fontdimen4\font\relax}
\providecommand\BIBforeignlanguage[2]{{%
\expandafter\ifx\csname l@#1\endcsname\relax
\typeout{** WARNING: IEEEtran.bst: No hyphenation pattern has been}%
\typeout{** loaded for the language `#1'. Using the pattern for}%
\typeout{** the default language instead.}%
\else
\language=\csname l@#1\endcsname
\fi
#2}}

\bibitem{geiger2012we}
A.~Geiger, P.~Lenz, and R.~Urtasun, ``Are we ready for autonomous driving? the
  kitti vision benchmark suite,'' in \emph{2012 IEEE conference on computer
  vision and pattern recognition}.\hskip 1em plus 0.5em minus 0.4em\relax IEEE,
  2012, pp. 3354--3361.

\bibitem{menze2015object}
M.~Menze and A.~Geiger, ``Object scene flow for autonomous vehicles,'' in
  \emph{Proceedings of the IEEE conference on computer vision and pattern
  recognition}, 2015, pp. 3061--3070.

\bibitem{yang2019drivingstereo}
G.~Yang, X.~Song, C.~Huang, Z.~Deng, J.~Shi, and B.~Zhou, ``Drivingstereo: A
  large-scale dataset for stereo matching in autonomous driving scenarios,'' in
  \emph{Proceedings of the IEEE/CVF Conference on Computer Vision and Pattern
  Recognition}, 2019, pp. 899--908.

\bibitem{huang2019apolloscape}
X.~Huang, P.~Wang, X.~Cheng, D.~Zhou, Q.~Geng, and R.~Yang, ``The apolloscape
  open dataset for autonomous driving and its application,'' \emph{IEEE
  transactions on pattern analysis and machine intelligence}, vol.~42, no.~10,
  pp. 2702--2719, 2019.

\bibitem{chang2019argoverse}
M.-F. Chang, J.~Lambert, P.~Sangkloy, J.~Singh, S.~Bak, A.~Hartnett, D.~Wang,
  P.~Carr, S.~Lucey, D.~Ramanan, \emph{et~al.}, ``Argoverse: 3d tracking and
  forecasting with rich maps,'' in \emph{Proceedings of the IEEE/CVF Conference
  on Computer Vision and Pattern Recognition}, 2019, pp. 8748--8757.

\bibitem{cordts2016cityscapes}
M.~Cordts, M.~Omran, S.~Ramos, T.~Rehfeld, M.~Enzweiler, R.~Benenson,
  U.~Franke, S.~Roth, and B.~Schiele, ``The cityscapes dataset for semantic
  urban scene understanding,'' in \emph{Proceedings of the IEEE conference on
  computer vision and pattern recognition}, 2016, pp. 3213--3223.

\bibitem{scharstein2002taxonomy}
D.~Scharstein and R.~Szeliski, ``A taxonomy and evaluation of dense two-frame
  stereo correspondence algorithms,'' \emph{International journal of computer
  vision}, vol.~47, no.~1, pp. 7--42, 2002.

\bibitem{scharstein2003high}
------, ``High-accuracy stereo depth maps using structured light,'' in
  \emph{2003 IEEE Computer Society Conference on Computer Vision and Pattern
  Recognition, 2003. Proceedings.}, vol.~1.\hskip 1em plus 0.5em minus
  0.4em\relax IEEE, 2003, pp. I--I.

\bibitem{scharstein2007learning}
D.~Scharstein and C.~Pal, ``Learning conditional random fields for stereo,'' in
  \emph{2007 IEEE Conference on Computer Vision and Pattern Recognition}.\hskip
  1em plus 0.5em minus 0.4em\relax IEEE, 2007, pp. 1--8.

\bibitem{hirschmuller2007evaluation}
H.~Hirschmuller and D.~Scharstein, ``Evaluation of cost functions for stereo
  matching,'' in \emph{2007 IEEE Conference on Computer Vision and Pattern
  Recognition}.\hskip 1em plus 0.5em minus 0.4em\relax IEEE, 2007, pp. 1--8.

\bibitem{scharstein2014high}
D.~Scharstein, H.~Hirschm{\"u}ller, Y.~Kitajima, G.~Krathwohl,
  N.~Ne{\v{s}}i{\'c}, X.~Wang, and P.~Westling, ``High-resolution stereo
  datasets with subpixel-accurate ground truth,'' in \emph{German conference on
  pattern recognition}.\hskip 1em plus 0.5em minus 0.4em\relax Springer, 2014,
  pp. 31--42.

\bibitem{schops2017multi}
T.~Schops, J.~L. Schonberger, S.~Galliani, T.~Sattler, K.~Schindler,
  M.~Pollefeys, and A.~Geiger, ``A multi-view stereo benchmark with
  high-resolution images and multi-camera videos,'' in \emph{Proceedings of the
  IEEE Conference on Computer Vision and Pattern Recognition}, 2017, pp.
  3260--3269.

\bibitem{bao2020instereo2k}
W.~Bao, W.~Wang, Y.~Xu, Y.~Guo, S.~Hong, and X.~Zhang, ``Instereo2k: a large
  real dataset for stereo matching in indoor scenes,'' \emph{Science China
  Information Sciences}, vol.~63, no.~11, pp. 1--11, 2020.

\bibitem{wang2019irs}
Q.~Wang, S.~Zheng, Q.~Yan, F.~Deng, K.~Zhao, and X.~Chu, ``Irs: A large
  naturalistic indoor robotics stereo dataset to train deep models for
  disparity and surface normal estimation,'' \emph{arXiv preprint
  arXiv:1912.09678}, 2019.

\bibitem{treible2017cats}
W.~Treible, P.~Saponaro, S.~Sorensen, A.~Kolagunda, M.~O'Neal, B.~Phelan,
  K.~Sherbondy, and C.~Kambhamettu, ``Cats: A color and thermal stereo
  benchmark,'' in \emph{Proceedings of the IEEE Conference on Computer Vision
  and Pattern Recognition}, 2017, pp. 2961--2969.

\bibitem{kendall2017end}
A.~Kendall, H.~Martirosyan, S.~Dasgupta, P.~Henry, R.~Kennedy, A.~Bachrach, and
  A.~Bry, ``End-to-end learning of geometry and context for deep stereo
  regression,'' in \emph{Proceedings of the IEEE International Conference on
  Computer Vision}, 2017, pp. 66--75.

\bibitem{chang2018pyramid}
J.-R. Chang and Y.-S. Chen, ``Pyramid stereo matching network,'' in
  \emph{Proceedings of the IEEE Conference on Computer Vision and Pattern
  Recognition}, 2018, pp. 5410--5418.

\bibitem{guo2019group}
X.~Guo, K.~Yang, W.~Yang, X.~Wang, and H.~Li, ``Group-wise correlation stereo
  network,'' in \emph{Proceedings of the IEEE/CVF Conference on Computer Vision
  and Pattern Recognition}, 2019, pp. 3273--3282.

\bibitem{li2021revisiting}
Z.~Li, X.~Liu, N.~Drenkow, A.~Ding, F.~X. Creighton, R.~H. Taylor, and
  M.~Unberath, ``Revisiting stereo depth estimation from a sequence-to-sequence
  perspective with transformers,'' in \emph{Proceedings of the IEEE/CVF
  International Conference on Computer Vision}, 2021, pp. 6197--6206.

\bibitem{mayer2016large}
N.~Mayer, E.~Ilg, P.~Hausser, P.~Fischer, D.~Cremers, A.~Dosovitskiy, and
  T.~Brox, ``A large dataset to train convolutional networks for disparity,
  optical flow, and scene flow estimation,'' in \emph{Proceedings of the IEEE
  conference on computer vision and pattern recognition}, 2016, pp. 4040--4048.

\bibitem{butler2012naturalistic}
D.~J. Butler, J.~Wulff, G.~B. Stanley, and M.~J. Black, ``A naturalistic open
  source movie for optical flow evaluation,'' in \emph{European conference on
  computer vision}.\hskip 1em plus 0.5em minus 0.4em\relax Springer, 2012, pp.
  611--625.

\bibitem{ladicky2012joint}
L.~Ladick{\`y}, P.~Sturgess, C.~Russell, S.~Sengupta, Y.~Bastanlar,
  W.~Clocksin, and P.~H. Torr, ``Joint optimization for object class
  segmentation and dense stereo reconstruction,'' \emph{International Journal
  of Computer Vision}, vol. 100, no.~2, pp. 122--133, 2012.

\bibitem{hirschmuller2007stereo}
H.~Hirschmuller, ``Stereo processing by semiglobal matching and mutual
  information,'' \emph{IEEE Transactions on pattern analysis and machine
  intelligence}, vol.~30, no.~2, pp. 328--341, 2007.

\bibitem{kolmogorov2001computing}
V.~Kolmogorov and R.~Zabih, ``Computing visual correspondence with occlusions
  using graph cuts,'' in \emph{Proceedings Eighth IEEE International Conference
  on Computer Vision. ICCV 2001}, vol.~2.\hskip 1em plus 0.5em minus
  0.4em\relax IEEE, 2001, pp. 508--515.

\bibitem{klaus2006segment}
A.~Klaus, M.~Sormann, and K.~Karner, ``Segment-based stereo matching using
  belief propagation and a self-adapting dissimilarity measure,'' in \emph{18th
  International Conference on Pattern Recognition (ICPR'06)}, vol.~3.\hskip 1em
  plus 0.5em minus 0.4em\relax IEEE, 2006, pp. 15--18.

\bibitem{he2021semi}
J.~He, E.~Zhou, L.~Sun, F.~Lei, C.~Liu, and W.~Sun, ``Semi-synthesis: A fast
  way to produce effective datasets for stereo matching,'' in \emph{Proceedings
  of the IEEE/CVF Conference on Computer Vision and Pattern Recognition}, 2021,
  pp. 2884--2893.

\bibitem{yang2019hierarchical}
G.~Yang, J.~Manela, M.~Happold, and D.~Ramanan, ``Hierarchical deep stereo
  matching on high-resolution images,'' in \emph{Proceedings of the IEEE/CVF
  Conference on Computer Vision and Pattern Recognition}, 2019, pp. 5515--5524.

\bibitem{peris2012towards}
M.~Peris, S.~Martull, A.~Maki, Y.~Ohkawa, and K.~Fukui, ``Towards a simulation
  driven stereo vision system,'' in \emph{Proceedings of the 21st International
  Conference on Pattern Recognition (ICPR2012)}.\hskip 1em plus 0.5em minus
  0.4em\relax IEEE, 2012, pp. 1038--1042.

\bibitem{wang2019flickr1024}
Y.~Wang, L.~Wang, J.~Yang, W.~An, and Y.~Guo, ``Flickr1024: A large-scale
  dataset for stereo image super-resolution,'' in \emph{Proceedings of the
  IEEE/CVF International Conference on Computer Vision Workshops}, 2019, pp.
  0--0.

\bibitem{hua2020holopix50k}
Y.~Hua, P.~Kohli, P.~Uplavikar, A.~Ravi, S.~Gunaseelan, J.~Orozco, and E.~Li,
  ``Holopix50k: A large-scale in-the-wild stereo image dataset,'' \emph{arXiv
  preprint arXiv:2003.11172}, 2020.

\bibitem{borstler2016role}
J.~B{\"o}rstler and B.~Paech, ``The role of method chains and comments in
  software readability and comprehension—an experiment,'' \emph{IEEE
  Transactions on Software Engineering}, vol.~42, no.~9, pp. 886--898, 2016.

\bibitem{poggi2021synergies}
M.~Poggi, F.~Tosi, K.~Batsos, P.~Mordohai, and S.~Mattoccia, ``On the synergies
  between machine learning and binocular stereo for depth estimation from
  images: a survey,'' \emph{IEEE Transactions on Pattern Analysis and Machine
  Intelligence}, 2021.

\bibitem{zbontar2015computing}
J.~Zbontar and Y.~LeCun, ``Computing the stereo matching cost with a
  convolutional neural network,'' in \emph{Proceedings of the IEEE conference
  on computer vision and pattern recognition}, 2015, pp. 1592--1599.

\bibitem{seki2017sgm}
A.~Seki and M.~Pollefeys, ``Sgm-nets: Semi-global matching with neural
  networks,'' in \emph{Proceedings of the IEEE Conference on Computer Vision
  and Pattern Recognition}, 2017, pp. 231--240.

\bibitem{gidaris2017detect}
S.~Gidaris and N.~Komodakis, ``Detect, replace, refine: Deep structured
  prediction for pixel wise labeling,'' in \emph{Proceedings of the IEEE
  conference on computer vision and pattern recognition}, 2017, pp. 5248--5257.

\bibitem{ranftl2021vision}
R.~Ranftl, A.~Bochkovskiy, and V.~Koltun, ``Vision transformers for dense
  prediction,'' in \emph{Proceedings of the IEEE/CVF International Conference
  on Computer Vision}, 2021, pp. 12\,179--12\,188.

\bibitem{wu2019semantic}
Z.~Wu, X.~Wu, X.~Zhang, S.~Wang, and L.~Ju, ``Semantic stereo matching with
  pyramid cost volumes,'' in \emph{Proceedings of the IEEE/CVF International
  Conference on Computer Vision}, 2019, pp. 7484--7493.

\bibitem{kusupati2020normal}
U.~Kusupati, S.~Cheng, R.~Chen, and H.~Su, ``Normal assisted stereo depth
  estimation,'' in \emph{Proceedings of the IEEE/CVF Conference on Computer
  Vision and Pattern Recognition}, 2020, pp. 2189--2199.

\bibitem{song2018edgestereo}
X.~Song, X.~Zhao, H.~Hu, and L.~Fang, ``Edgestereo: A context integrated
  residual pyramid network for stereo matching,'' in \emph{Asian Conference on
  Computer Vision}.\hskip 1em plus 0.5em minus 0.4em\relax Springer, 2018, pp.
  20--35.

\bibitem{zhang2000flexible}
Z.~Zhang, ``A flexible new technique for camera calibration,'' \emph{IEEE
  Transactions on pattern analysis and machine intelligence}, vol.~22, no.~11,
  pp. 1330--1334, 2000.

\end{thebibliography}
	
\end{document}